\documentclass[10pt,twocolumn,letterpaper]{article}

\usepackage{iccv}
\usepackage{times}
\usepackage{epsfig}
\usepackage{graphicx}
\usepackage{amsmath}
\usepackage{amssymb}
\usepackage{multirow}
\usepackage{tabularx}
\usepackage{booktabs}

\usepackage[breaklinks=true,bookmarks=false]{hyperref}

\iccvfinalcopy 

\def\iccvPaperID{****} 

\ificcvfinal\fi

\begin{document}

\title{Contrastive Image Synthesis and Self-supervised Feature Adaptation for Cross-Modality Biomedical Image Segmentation}

\def\authorBlock{
    Xinrong Hu$^{1}$ \qquad
    Corey Wang$^{2}$ \qquad
    Yiyu Shi$^{1}$ \\
    $^{1}$ Department of Computer Science and Engineering, University of Notre Dame \\
     $^{2}$ Northwestern University \\
    {\tt\small \{xhu7, yshi4\}@nd.edu}
}
\author{\authorBlock}
\maketitle
\ificcvfinal\thispagestyle{empty}\fi

\begin{abstract}
This work presents a novel framework \emph{CISFA} (\textbf{C}ontrastive \textbf{I}mage synthesis and \textbf{S}elf-supervised \textbf{F}eature \textbf{A}daptation) that builds on image domain translation and unsupervised feature adaptation for cross-modality biomedical image segmentation. 
Different from existing approaches, our method employs a one-sided generative model and incorporates a weighted patch-wise contrastive loss between sampled patches of the input image and the corresponding synthetic image, which serves as shape constraints.
Furthermore, we notice that the generated images and input images share similar structural information but are in different modalities. 
To address this, we enforce contrastive losses on the generated images and the input images to train the encoder of a segmentation model to minimize the discrepancy between paired images in the learned embedding space. Compared with existing works that rely on adversarial learning for feature adaptation, such a method enables the encoder to learn domain-independent features in a more explicit way. 
We extensively evaluate our methods on segmentation tasks containing CT and MRI images 
for abdominal cavities and whole hearts.
Experimental results show that the proposed framework not only outputs synthetic images with less distortion of organ shapes, but also outperforms state-of-the-art domain adaptation methods. Codes are available in \href{https://github.com/xhu248/cisfa_da}{https://github.com/xhu248/cisfa\_da}.
\end{abstract}

\section{Introduction}

Due to the nature of supervised learning, the performance of deep neural networks (DNN) suffers from severe degradation when the domain distribution shifts \cite{wang2018deep, csurka2017domain, wilson2020survey}.   
One common scenario where such shifts occur is among 
biomedical images. 
In clinical practice, computed tomography (CT) and magnetic resonance imaging (MRI) are two common medical radiological imaging techniques. 
The distinct imaging mechanisms result in dissimilarity of CT and MRI with respect to brightness, contrast, and texture. 
A deep learning model trained on CT images to segment brain tumours may fail to achieve comparable accuracy on a brain MRI scan.
Besides modality, the difference between CT scanners made by different manufacturers can also cause the accuracy to decrease.
However, it is infeasible to collect labeled datasets for all possible biomedical image domains, as pixel-wise labelling is laborious and requires expert knowledge, not to mention the privacy issues.

All the above obstacles give rise to unsupervised domain adaptation (UDA) study.
There have been numerous works aiming at improving semantic segmentation performance on target domain with only the source domain annotated. 
One branch of these approaches is to train a single model that is capable of segmenting different styled images, and the key is to extract common features shared by different domains through the encoding path \cite{tsai2018learning}. 
However, this kind of feature adaptation is coarse-grained. 
Images with different structural information are forced to be similar in the embedding space, and there could still be a distribution margin between two domains that the discriminator fails to detect.
In contrast, \cite{zhang2019category} applied the L2 norm of the difference between features and the category anchors to drive the intra-category features closer regardless of domains.
However, they assumed that a model pretrained on the source domain can generate reliable pseudo-label for target domain images, which is not necessarily true especially when the domain shift is dramatic. 

Since the unpaired image translation problem was well tackled by CycleGAN\cite{zhu2017unpaired}, taking advantage of generative models has become another stream in the field of UDA segmentation. 
The widely used strategy is integrating a segmentation model with the CycleGAN framework.
After the source domain images are transferred to the target domain, the corresponding labels would supervise the segmentation training in the target domain \cite{hoffman2018cycada, huo2018adversarial, chen2020unsupervised, tomar2021self}. 
One drawback of CycleGAN based approaches is too many models in the overall framework, especially as some works add additional discriminators and encoding models for feature adaptation. The large number of models makes it difficult to optimize the parameters and lengthens training iterations. 
Moreover, although there is an identity loss function in CycleGAN that drives the reconstructed image to be exactly the same as the input image, there is no direct constraint on the input image and the translated image to avoid spatial distortion or structure variation.

\begin{figure*}
    \centering
    \includegraphics[width=0.8\linewidth]{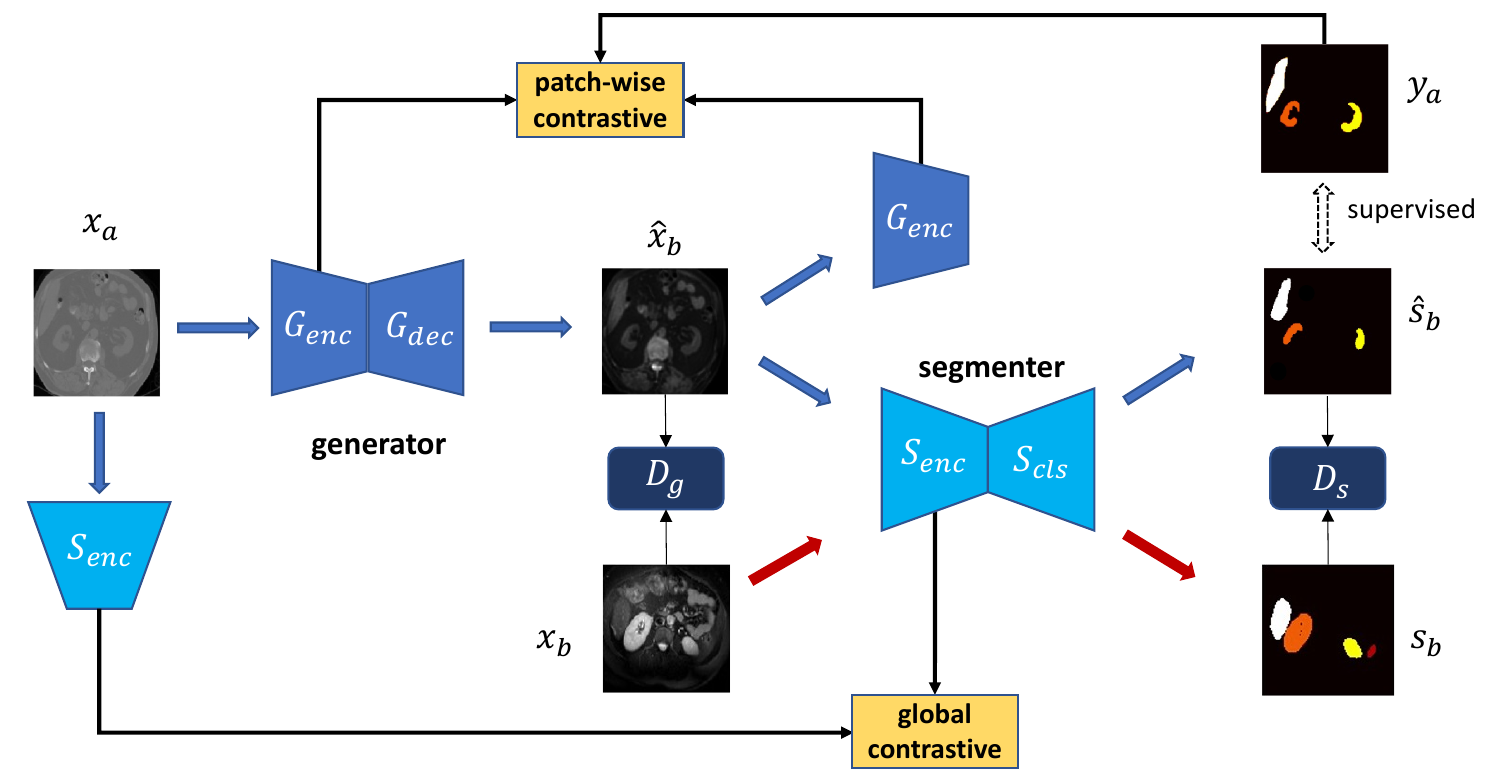}
    \caption{Overview of \emph{CISFA} framework. The main components include a generator $G$, a segmenter $Seg$, and two discriminators, $D_{g}$ \& $D_{s}$. There are two contrastive losses: a global contrastive loss for self-supervised feature adaptation, and a patch-wise contrastive loss for keeping the shape consistency in synthetic images. Blue/red arrows represent the data flow for source/target domain images respectively.}
    \label{fig1}
\end{figure*}

Being aware of previous works' limitations, this paper proposes a novel framework \emph{CISFA} featuring  lighter generative models and self-supervised feature adaptation, comprising of a generator for image synthesis and a segmenter for target domain.
For image synthesis, inspired by a new image translation model named CUT\cite{park2020contrastive}, we remove the path that translates target domain to source domain in the CycleGAN flow to facilitate training.
To exert the shape-consistency constraint, we maximize the mutual information for image patches at the same position between input images and translated images in latent space.
We further extend the pixel-wise InfoNCE loss in CUT by introducing an attention mechanism generated from segmentation labels.
This is done by adding weights to the contrastive loss for non-background pixels so that the area of interest is emphasized more during translation.
On the other hand, ideally the segmenter only sees target domain images, but there inevitably exists gap between the distribution of generated dataset and real objective dataset. 
Consequently, feature adaptation is still beneficial for the segmenter.
Observing that a successful generator outputs an image that has exactly the same contents and details as the input image except for modality, we realize these paired images serve as good examples of a positive pair in \cite{chen2020simple}.
Thus, we decide to make use of self-supervised learning to conduct feature level adaptation for the segmentation model, different from all previous works. 
In detail, we add a multi-layer perceptron to the encoder of the segmenter, and project input images as well as synthetic images to one-dimensional features.
Then the encoder learns domain invariant features by reducing the cosine distance of paired features.

We conduct experiments with our method and the state-of-the-arts on two medical image UDA tasks. 
For the first task, we collect 20 CT and 30 MRI abdomen scans from two public datasets, and for the second, we use the MWWHS dataset that contains 20 CT and 20 MRI whole heart 3D images. 
According to the experiment results, \emph{CISFA} demonstrates superior performance in terms of segmentation accuracy on the target domain than existing works. 

\section{Related Works}
\label{sec:related}
\textbf{Domain Adaption for Semantic Segmentation}
One straightforward solution to this problem is training the model to learn domain-independent features. 
\cite{luo2019taking} added two classifiers to the encoder in order to get diverse view for a feature, and they extended traditional adversarial loss with an adaptive loss, which was calculated by the cosine distance between the output of those two classifiers. 
Therefore, the follow-up discriminator would focus more on poorly aligned categories.
Instead of using adversarial learning, \cite{zhang2019category} tried to aggregate features belonging to the same categories for different domain images. Since they did not have annotation for the target domain images, they used a model pretrained on the source domain to generate pseudo-labels.
\cite{ouyang2019data} fused prior matching to a VAE model to learn a shared feature space between two domains. 
More specifically, two domains had a shared VAE module and distinct encoder and decoder modules, and they applied adversarial learning to align features extracted from different encoders.
The work\cite{wu2021unsupervised} was also based on VAE, 
and they drove the distribution of feature maps to the same parameterized variational form. 

Image translation is an alternative way to tackle UDA. Compared to feature alignment in latent space, we can just generate images out of an existing dataset that shares a similar distribution as another dataset. After that, we can take advantage of the annotated domain to do supervised training.
The feasibility is ascribed to the rapid development of generative adversarial model (GAN), especially the emergence of CycleGAN\cite{zhu2017unpaired} in unpaired image translation. 
\cite{hoffman2018cycada} firstly integrated a segmentation model with the CycleGAN module for road scene images from different sources.
At the same time, \cite{huo2018adversarial} used the integrated framework to achieve domain adaption on CT-MR abdomen segmentation.
SIFA\cite{chen2020unsupervised} further extended the workflow by making the segmenter and the source domain generators share the encoder, adding a discriminator to the segmentation results as well as the latent embedding in generated image space.
Most recently, \cite{tomar2021self} added an attention mechanism in the form of normalization to features in the generator to improve the quality of synthesized images for SIFA.

\textbf{Contrastive Learning}
Recently, contrastive learning, as a self-supervised learning method, gained popularity after the work SimCLR\cite{chen2020simple}, in which, original image and its augmented views are clustered in feature space.
This pretext task can learn useful representations that can boost downstream tasks, like image classification and object detection.
Follow-up works provided different perspectives of influential factors, such as transformation combinations\cite{chen2020big}, batch size and momentum encoders\cite{he2020momentum, chen2021exploring}, as well as supervised class clustering\cite{khosla2020supervised}. 
Due to the powerful ability of representation learning, this emergent technique has been widely applied to medical image domain, including classification\cite{gazda2021self, ciga2022self, li2021dual, he2020sample} and segmentation task\cite{chaitanya2020contrastive, zeng2021positional, taleb20203d, zhang2022unsupervised}.
For example, \cite{chaitanya2020contrastive, zeng2021positional} observed the similarity of adjacent slices in 3D medical volumes and explored how this kind of knowledge could improve the segmentation performance with less labeled data.
However, there are few works that utilize contrastive loss for unsupervised domain adaptation.
\section{Methods}
\label{sec:methods}

\subsection{Overview}

Firstly, we will formulate the UDA segmentation problem for medical imaging. 
Given two datasets, $A = \left \{ (x_{a}, y_{a}) | x_{a}  \in \mathcal{A}  \right \}$ with labels and $B = \left \{ x_{b} \in \mathcal{B} \right \}$ without labels, we aim to generate segmentation masks $\hat{y}_{b}$ for images in $B$. $\mathcal{A}$ and $\mathcal{B}$ represent the source domain and the target domain respectively, and can be different modalities or be collected from different scanners.
Since labels are missing in the target domain, supervised learning is not applicable in this situation. 
Moreover, as there exists a distribution shift from $\mathcal{A}$ to $\mathcal{B}$, models trained on source dataset fails to give satisfying segmentation accuracy for images in $B$. 

Fig.\ref{fig1} shows the sketch of the overall framework CISFA to tackle the above UDA problem.
We use $G(\cdot)$ to translate $x_{a} \rightarrow \hat{x}_{b} = G(x_{a}) \in \hat{\mathcal{B}}$ while maintaining the structural contents in $x_{a}$. 
Then we get labeled data $(\hat{x}_{b}, y_{a})$ to supervise the training of a segmentation model $Seg(\cdot)$.
CISFA is different from existing frameworks in four aspects: 1) 
We use one fewer generative model than CycleGAN based models. This makes the model easier to train. 2) We use two different contrastive losses to deal with structure distortion and domain shift. 3) Our patch-wise contrastive loss, unlike the losses used in the literature for the same purpose, assigns different weights for different patches, enforcing more attention to non-background patches. 4) Unlike existing works that use adversarial learning for feature adaptation, we introduce a global contrastive loss for model to learn domain invariant features.
In the rest of this section, we will firstly introduce the detail of the image translation process and then the patch-wise contrastive loss. 
Subsequently, we will describe the self-supervised feature adaptation for the downsampling path of $Seg$ as well as other losses regarding the segmenter $Seg$.
Lastly, we will present some training details of all models in the framework.

\begin{figure*}
    \centering
    \includegraphics[width=0.9\linewidth]{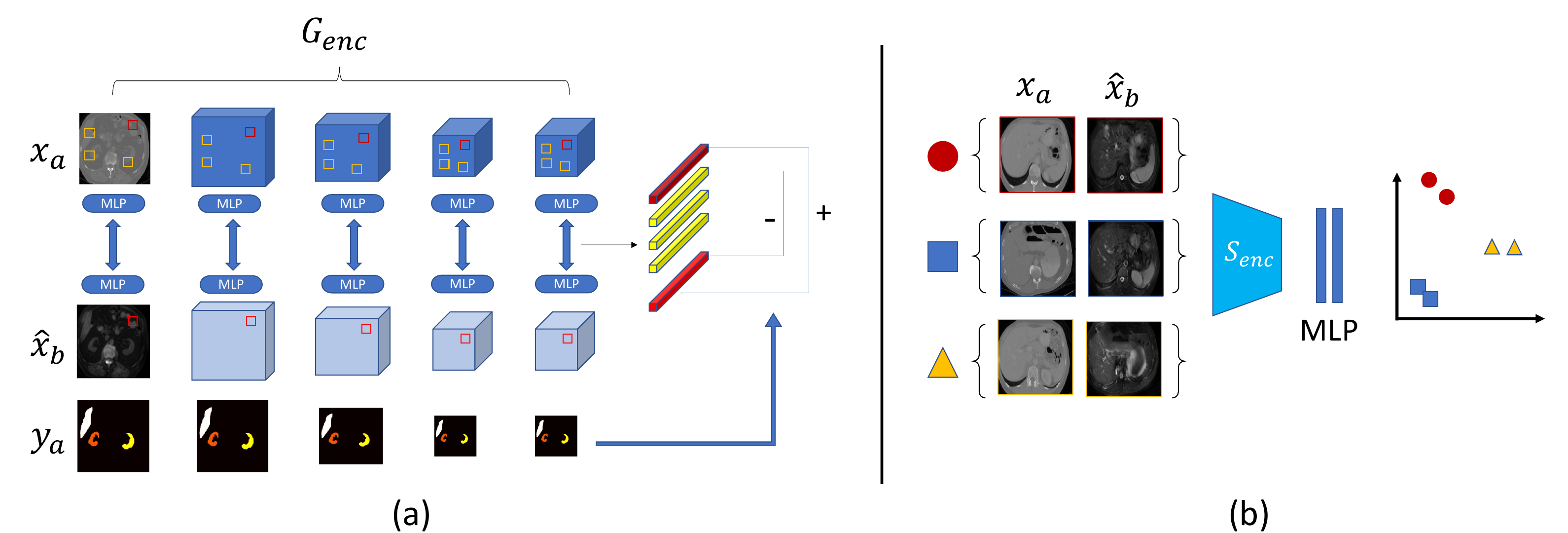}
    \caption{Illustration of the calculation of (a) the patch-wise contrastive loss and (b) global contrastive loss. (a): We select four layers inside the $G_{enc}$ as well as the original images. The small frame in the feature map represents randomly sampled patches. "+" stands for positive pairs and "-" stands for negative pairs. We also downsample the label to the same resolution as each layer of feature maps and increase the weights for non-background patches. (b): Patterns with the same shape and the same color denote positive pairs and should be closer in the latent space. }
    \label{fig2}
\end{figure*}

\subsection{Weighted Patch-wise Contrastive Loss for Image Synthesis}
Similar to all generative adversarial networks, we add a discriminator $D_{g}(\cdot)$ to distinguish between the  target domain image $x_{a}$ and fake target domain image $\hat{x}_{b}$. 
This is done by minimizing the following loss,
\begin{equation}
\begin{split}
    L_{D\_G} =& \mathbb{E}_{x_{b}\sim B}[logD_{g}(x_{b})] + \\
     & \mathbb{E}_{x_{a}\sim A}[log(1 - D_{g}(G(x_{a}))]
\end{split}
\end{equation}
Simultaneously, the task of $G$ is to deceive the discriminator into classifying synthetic images as in real $\mathcal{B}$.
Gradually, $G$ learns to generate images sharing the similar texture as $x_{b}$.  

The consistency of structural information between $x_{a}$ and $\hat{x}_{b}$ assures that the annotation is still correct for the after-translation image, which is the key to the success of the following supervised training.
However, lacking the path that transforms $\hat{x}_{b}$ back to domain $\mathcal{A}$, \emph{CISFA} has no cycle loss that penalizes the distortion after image translation.
Therefore, we devise a patch-wise contrastive loss as an alternative shape consistency constraint.
Specifically, we extract feature maps of input images and output images in different levels from the encoder of the generative model $G_{enc}$, $\left \{ f_{a}^{l} | l = 1,2,..,\right \} = G_{enc}(x_{a})$, $\left \{ f_{\hat{b}}^{l} | l = 1,2,..,\right \} = G_{enc}(\hat{x}_{b})$. 
For feature map $f_{a}^{l} \in \mathbb{R}^{c*h*w}$, where $c$ is the channel number, $h*w$ is the feature map size, $l$ denotes the $lth$ layer. 
Taking all features into account is computationally expensive. Hence we randomly sample features in dimension $h*w$ for $f_{a}^{l}$ and do sampling from the same position for  $f_{\hat{a}}^{l}$.
Afterwards, we add a multi-layer perceptron (MLP) as projection head to the sampled features and apply $L2$ normalization. Then we get a feature vector set $F^{l}=\left\{\mathbf{u} | \mathbf{u} \in \mathbb{R}^{c'}\right \}$, where $c'$ is the feature dimension and is set as 128.
Let $\mathbf{u^{+}} \in F^{l}$ denote the feature vector derived from the same position as $u$ in the other feature map. 
The patch-wise contrastive loss for layer $l$ is defined as
\begin{equation}
    L_{pcl}(l, u) = -log\frac{exp(\mathbf{u} \cdot \mathbf{u^{+}}/\tau )}{\sum_{v \in F^{l} - u} exp(\mathbf{u} \cdot \mathbf{v}/\tau)}
\end{equation}
in which, $\tau$ is the temperature parameter for the cosine difference and we choose 0.2 for $\tau$ in this work.
This loss reduces the discrepancy between the same patch before and after transformation in the latent space.

Being aware of the fact that the label $y_{a}$ is also provided, in this work, we extend the definition of patch-wise contrastive loss by adding weights to features corresponding to different categories.
Since the translation correctness of non-background areas is more important to the consequential segmentation training compared to distortion of background pixels, we set the weights of non-background features to be larger than background ones.
In practice, for each layer $l$, we downsample the label to the same resolution as the feature map $f^{l}$ and decide the weight for each sampled feature vector. 
The newly added weight $w_p(u)$ is 

\begin{equation}
     w_{p}(u) = \left\{\begin{matrix} 1, & \mbox{if u is background} \\ 
    w, & \mbox{else}
    \end{matrix}\right.
\end{equation}
where w can be any number larger than 1 and we set it as 2 in this work.
Then the total loss for generator $G$ is,

\begin{equation}
\begin{split}
    L_{G} = & \mathbb{E}_{x_{a}\sim A}[logD_{g}(G(x_{a})] + \\
    & \mathbb{E}_{x_{a}\sim A,l\sim L, u~\sim F_{l}}[w_{p}(u)*L_{pcl}(l, u)]
\end{split}
\end{equation}
in which $L$ is the set of selected layers to calculate patch-wise contrastive loss.

\subsection{Self-supervised Feature Adaptation}
After obtaining the translated image $\hat{x}_{b}$, we pass it into the segmentation model $Seg$ to get a prediction mask $\hat{s}_b$. 
By optimizing the dice loss of $(\hat{s}_b, y_{a})$, we expect the segmenter to learn how to tackle images in the target domain.
As shown in previous works \cite{chaitanya2020contrastive, zeng2021positional}, pretraining the encoder path of segmentation model to learn mutual information among similar slices benefits the follow-up segmentation task, where similar slices refer to images at very adjacent positions in the volume.
For our case, we observe that $x_{a}$ and $\hat{x}_{b}$ share the same content but differ in modality, which are good examples of positive pairs in contrastive learning.
Let $S_{enc}(\cdot)$ denote the downsampling path of the segmenter.
We project $t$ pairs of input images and their corresponding synthetic images into latent space, and add a MLP head for stronger representation ability in the feature space.
After normalization, we get features $\left \{ z_{i} = \left \|  \mbox{MLP}(S_{enc}(x^{i})) \right \| | i = 1, 2, ..., 2t\right \}$, and assume that $j(i)$ is the index of positive pair feature regarding $z_{i}$.
We then calculate the contrastive loss similar to SimCLR\cite{chen2020simple}.
\begin{equation}
    L_{gcl} (t) =  -\frac{1}{2t}\sum_{i}log\frac{exp(z_{i}\cdot z_{j(i)}/\tau)}{\sum_{k \neq i}^{2t}exp(z_{i}\cdot z_{k}/\tau)},
\end{equation}
which is defined as global contrastive loss in this paper. 
In contrast to $L_{ncl}$, the feature vector $z_{i}$ contains the global information of the input image instead of only one patch.
Moreover, we also generate the prediction mask $s_{b} = Seg(x_{b})$ in the real target domain, and add a discriminator $D_{s}$ to identify the output of images in $\mathcal{B}$.
In that case, the segmenter is trained to give segmentation result of the same quality for $\hat{x}_{b}$ and $x_{b}$, even though there is still a minor difference in the distribution of $\hat{\mathcal{B}}$ and $\mathcal{B}$.
Then the overall loss for the $Seg$ is
\begin{equation}
\begin{split}
    L_{seg} = & \mathbb{E}_{x_{a} \sim A}[1 - Dice(Seg(G(x_{a}), y_a)] + \\
    & \mathbb{E}_{t \sim T}[L_{gcl}(t)] + \mathbb{E}_{x_{a} \sim A}[logD_{s}(\hat{s}_{b})]
\end{split}
\end{equation}

where T is the set of batches of paired input and synthetic images.
The loss that $D_{s}$ tries to minimize is formulated as
\begin{equation}
\begin{split}
    L_{D\_S} =& \mathbb{E}_{x_{b}\sim B}[logD_{s}(s_{b})] + \\
     & \mathbb{E}_{x_{a}\sim A}[log(1 - D_{s}(\hat{s}_{b}))]
\end{split}
\end{equation}

\subsection{Training Strategies}
Although we already cut one generative path in our framework compared to CycleGAN, there are still four models for training including $G$, $Seg$, $D_{g}$ and $D_{s}$.
We integrate all of them into one framework seamlessly and can train all the parameters end-to-end.
However, this does not mean we simply calculate all the loss functions and do the backpropagation at the same time. 
At each training iteration, the order of updating weights is actually $G \rightarrow Seg \rightarrow D_{g} \& D_{s}$.
Notice that, after changing weights of $G$, we do an inference on $G$ with the latest weights to update $\hat{x}_{b}$ and then optimize parameters for $Seg$.
Similarly, we get the prediction masks $s_{b}$ and $\hat{s}_{b}$ with the fresh $Seg$ before changing parameters for the two discriminators.
Therefore, in the next iteration, when calculating the generative loss for $G$ or $Seg$, the corresponding classifier has seen the new images or segmentation masks, which is then a fair game for the two min-max game players.
After training, we then obtain a segmenter that is capable of making pixel-wise prediction on the target domain without a single label.

\section{Experiment}
\label{sec:exp}
\newcolumntype{Y}{>{\centering\arraybackslash}X}
\begin{table*}[t]
    \small
    \centering
    \caption{Comparison between state-of-the-art methods and the proposed methods w.r.t. segmentation dice scores on abdominal MRI volumes. The translation direction is CT $\rightarrow$ MRI. The average dice score and corresponding standard deviation over four independent folds are presented for all four organs, including liver, LK (left kidney, RK(right kidney) and spleen.  }
    \label{table1}
    \begin{tabularx}{\linewidth}{c|Y|Y|Y|Y|Y}
    \toprule
\multicolumn{1}{c|}{\multirow{2}{*}{Methods}}  & \multicolumn{5}{c}{Dice\% $\uparrow$}  \\ \cline{2-6}
 \multicolumn{1}{c|}{} &\multicolumn{1}{c|}{liver} & \multicolumn{1}{c|}{LK} & \multicolumn{1}{c|}{RK} & \multicolumn{1}{c|}{spleen}  & \multicolumn{1}{c}{avg}  \\
\hline
Supervised      &\multicolumn{1}{c|}{89.00$\pm$1.08} & \multicolumn{1}{c|}{87.19$\pm$2.49}  & \multicolumn{1}{c|}{83.31$\pm$5.05} &\multicolumn{1}{c|}{88.08$\pm$1.82}    &\multicolumn{1}{c}{86.90$\pm$2.19}    \\
W/o adaptation      &\multicolumn{1}{c|}{10.15$\pm$3.94} & \multicolumn{1}{c|}{3.67$\pm$3.57}  & \multicolumn{1}{c|}{4.04$\pm$2.95} &\multicolumn{1}{c|}{7.15$\pm$6.81} &\multicolumn{1}{c}{6.25$\pm$1.26}               \\
\hline
CUT\cite{park2020contrastive} &\multicolumn{1}{c|}{38.17$\pm$6.33} & \multicolumn{1}{c|}{32.20$\pm$10.69}  & \multicolumn{1}{c|}{34.01$\pm$9.32} &\multicolumn{1}{c|}{35.83$\pm$10.44} &\multicolumn{1}{c}{35.05$\pm$8.19}  \\
VarDA\cite{wu2021unsupervised} &\multicolumn{1}{c|}{41.63$\pm$1.77} & \multicolumn{1}{c|}{32.95$\pm$6.47}  & \multicolumn{1}{c|}{34.53$\pm$4.14} &\multicolumn{1}{c|}{32.23$\pm$4.72} &\multicolumn{1}{c}{35.33$\pm$2.60} \\ 
SASAN\cite{tomar2021self} &\multicolumn{1}{c|}{67.23$\pm$9.98} & \multicolumn{1}{c|}{61.41$\pm$12.95}  & \multicolumn{1}{c|}{67.94$\pm$14.63} &\multicolumn{1}{c|}{62.63$\pm$13.65} &\multicolumn{1}{c}{64.80$\pm$11.48}    \\
SIFA\cite{chen2020unsupervised} &\multicolumn{1}{c|}{77.24$\pm$2.03} & \multicolumn{1}{c|}{68.03$\pm$5.60}  & \multicolumn{1}{c|}{68.99$\pm$5.16} &\multicolumn{1}{c|}{66.79$\pm$4.87} &\multicolumn{1}{c}{70.26$\pm$3.69} \\ \hline
CISFA(no weight) &\multicolumn{1}{c|}{76.14$\pm$10.72} & \multicolumn{1}{c|}{72.12$\pm$4.52}  & \multicolumn{1}{c|}{\textbf{74.94$\pm$4.14}} &\multicolumn{1}{c|}{73.18$\pm$3.11} &\multicolumn{1}{c}{74.10$\pm$1.84} \\
CISFA & \textbf{80.13$\pm$2.21}  &\textbf{74.45$\pm$5.67} &74.51$\pm$5.16 & \textbf{75.86$\pm$5.28} &\textbf{76.24$\pm$2.17} \\

\bottomrule
\end{tabularx}
    
\end{table*}

\subsection{Dataset}
\textbf{Abdominal Dataset} 
This dataset contains 30 volumes of CT scans from the \emph{Multi-Atlas Labeling Beyond the Cranial Vault Challenge} \cite{multi-atlas} and 20 volumes of T2-SPIR MRI from the \emph{ISBI 2019 CHAOS CHALLENGE} \cite{kavur2021chaos}.
We choose four organs that are manually annotated on both datasets as the segmentation task, which are liver, right kidney, left kidney, and spleen.
After trimming the whole volume to only contain the region of interest (ROI), we reshape all slices in the transverse plane to be unified 196*196 by interpolation. 

\textbf{MMWHS Dataset}
\cite{zhuang2016multi} provides 20 CT and 20 MRI 3D cardiac images from different patients with annotations by expert radiologists for both modalities, and we focus on 4 anatomical structures as the segmentation objects, including ascending aorta, left atrium blood cavity, left ventricle blood cavity, and myocardium.
We also do the crop on the 3D volumes and reshape each images in the coronal view into the size of 160*160.

For preprocessing, we do normalization on cropped images for both datasets so that all pixels in a volume are zero mean and unit variance.
We split the two datasets into four folds on volume basis for cross-validation, and then decompose 3D volumes to 2D slices in every fold.
This ensures that slices from the same scan can only exist in the same folder. In the training only labels in the source domain are used. 
For example, if MRI is the target domain for an experiment,  we use three folds of CT and MRI slices for training and only CT labels are considered. 
The remaining CT fold is treated as validation set.
After training, we evaluate the segmenter on the remaining MRI fold, from which we compute the dice score as well as average symmetric surface distance (ASSD) for each volume, then report the average and the standard deviation for the fourfold runs.

\subsection{Settings}

The backbone of our generative model is based on ResNet\cite{he2016deep}. 
We firstly use two convolutional layers with stride equal to 2 to downsample the input image, followed by 9 residual blocks. 
As for the segmenter, we deploy a U-Net\cite{ronneberger2015u} with 4 resolution stages.
To reduce memory usage, we build a fully convolutional network with 3 layers as our discriminator architecture, same as that described in \cite{isola2017image}. 
We implement the generative models and discriminator with the deep learning package PyTorch. 
The GPU devices used are two NVIDIA Tesla P100 with 16GB memory each.
The optimizers used for updating weights are all based on the Adam\cite{kingma2014adam} algorithm.
The learning rate is also the same for all four models, lr=0.0002, $(\beta_{1}, \beta_{2}) = (0.5, 0.999)$.
The batch size is set as 4, considering the limitation of memory.
The training iteration number is 200 because we observe convergence of losses for all models after training for that number of epochs.

\begin{table*}[t]
    \small
    \centering
    \caption{Comparison between state-of-the-art methods and the proposed methods w.r.t. segmentation dice scores on abdominal CT volumes. The translation direction is MRI $\rightarrow$ CT.}
    \label{table3}
    \begin{tabularx}{\linewidth}{c|Y|Y|Y|Y|Y}
    \toprule
\multicolumn{1}{c|}{\multirow{2}{*}{Methods}}  & \multicolumn{5}{c}{Dice\% $\uparrow$}  \\ \cline{2-6}
 \multicolumn{1}{c|}{} &\multicolumn{1}{c|}{liver} & \multicolumn{1}{c|}{LK} & \multicolumn{1}{c|}{RK} & \multicolumn{1}{c|}{spleen}  & \multicolumn{1}{c}{avg}  \\
\hline
Supervised      &\multicolumn{1}{c|}{89.03$\pm$.95} & \multicolumn{1}{c|}{85.53$\pm$12.79}  & \multicolumn{1}{c|}{83.94$\pm$9.46} &\multicolumn{1}{c|}{85.49$\pm$4.05}    &\multicolumn{1}{c}{86.00$\pm$3.67}    \\
W/o adaptation      &\multicolumn{1}{c|}{9.38$\pm$3.08} & \multicolumn{1}{c|}{8.88$\pm$1.26}  & \multicolumn{1}{c|}{8.40$\pm$1.31} &\multicolumn{1}{c|}{9.70$\pm$1.52} &\multicolumn{1}{c}{9.09$\pm$0.68}               \\
\hline
CUT\cite{park2020contrastive} &\multicolumn{1}{c|}{17.78$\pm$8.74} & \multicolumn{1}{c|}{28.34$\pm$8.05}  & \multicolumn{1}{c|}{21.16$\pm$11.83} &\multicolumn{1}{c|}{19.29$\pm$10.60} &\multicolumn{1}{c}{21.64$\pm$8.66}  \\
VarDA\cite{wu2021unsupervised} &\multicolumn{1}{c|}{32.78$\pm$2.29} & \multicolumn{1}{c|}{38.11$\pm$4.17}  & \multicolumn{1}{c|}{31.71$\pm$4.32} &\multicolumn{1}{c|}{30.26$\pm$3.33} &\multicolumn{1}{c}{33.22$\pm$2.38} \\ 
SASAN\cite{tomar2021self}      &\multicolumn{1}{c|}{75.36$\pm$4.24} & \multicolumn{1}{c|}{67.33$\pm$6.43}  & \multicolumn{1}{c|}{67.25$\pm$6.08} &\multicolumn{1}{c|}{58.70$\pm$15.24} &\multicolumn{1}{c}{67.13$\pm$4.32}    \\
SIFA\cite{chen2020unsupervised} &\multicolumn{1}{c|}{74.03$\pm$1.13} & \multicolumn{1}{c|}{65.21$\pm$9.88}  & \multicolumn{1}{c|}{63.17$\pm$10.91} &\multicolumn{1}{c|}{63.53$\pm$11.85} &\multicolumn{1}{c}{66.49$\pm$5.61} \\ \hline
CISFA(no weight) &\multicolumn{1}{c|}{\textbf{77.45$\pm$2.15}} & \multicolumn{1}{c|}{66.91$\pm$7.16}  & \multicolumn{1}{c|}{64.92$\pm$4.57} &\multicolumn{1}{c|}{65.40$\pm$13.12} &\multicolumn{1}{c}{68.67$\pm$2.03} \\
CISFA &75.78$\pm$3.70 &\textbf{69.30$\pm$7.77} &\textbf{70.15$\pm$4.77} &\textbf{66.57$\pm$12.40} &\textbf{70.45$\pm$2.81} \\

\bottomrule
\end{tabularx}
    
\end{table*}

\begin{figure*}
    \centering
    \includegraphics[width=0.8\linewidth]{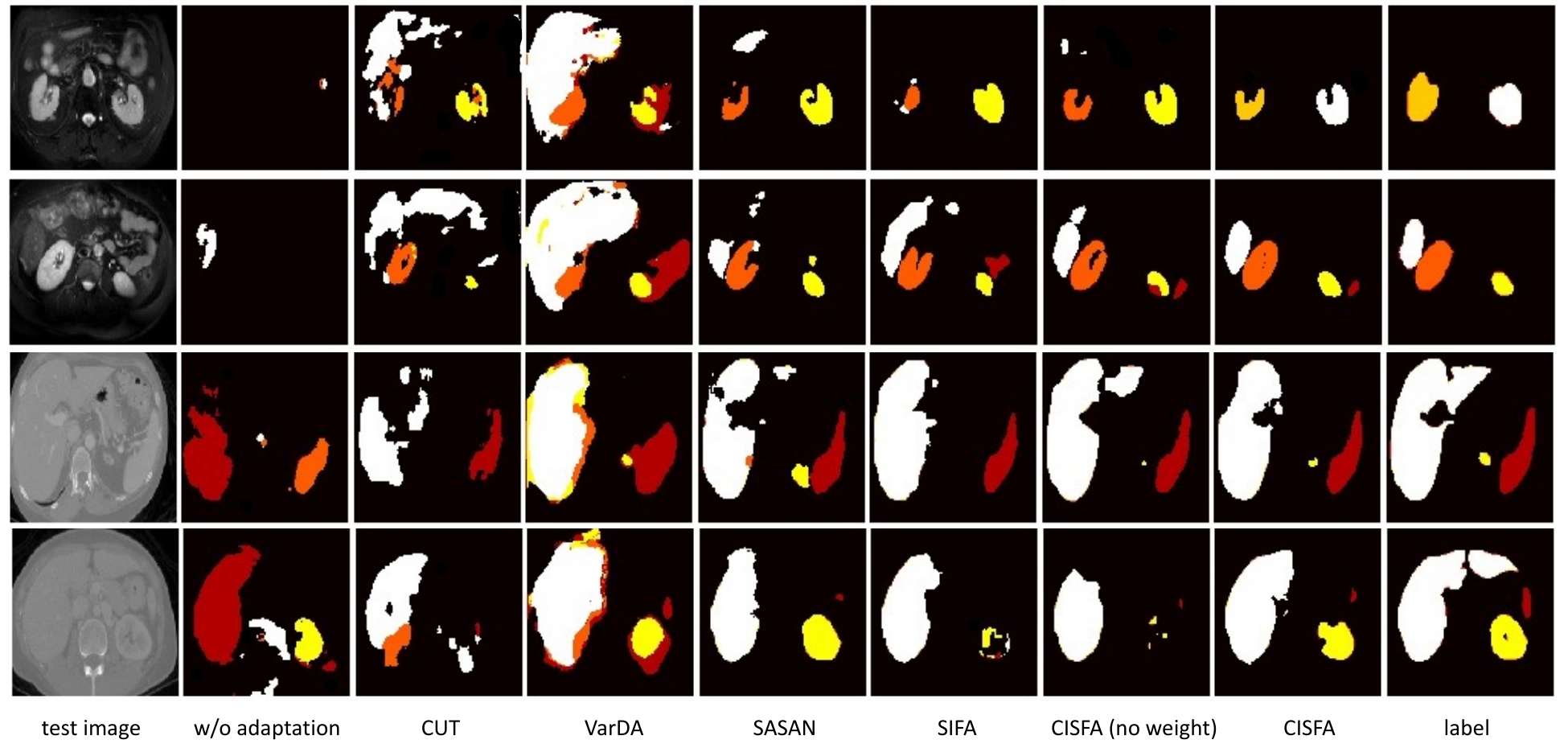}
    \caption{Visualization of segmentation results on the target domain regarding various methods. For the first two rows , the domain translation direction is CT $\rightarrow$ MRI, and for the other two rows, the direction is MRI $\rightarrow$ CT. }
    \label{fig3}
\end{figure*}

\subsection{Baselines}
To validate the effectiveness of our proposed method,
we conduct experiments with state-of-the-art approaches under the same setting as comparison.
CUT\cite{park2020contrastive} is a representative for image translation methods, and the segmentation part is not trained at the same time as the generative model. 
Accordingly, we just use CUT framework to translate images from source domain to target domain, and use the fake target images with source domain labels to supervise the training of the segmentation model.  
The reason why we don not include the results of CycleGAN is that the baseline methods below already shows their methods have superior peformance than CycleGAN, but none of them gives the comparison with CUT.
VarDA \cite{wu2021unsupervised} is the latest work that only uses feature adaptation in the field of biomedical UDA segmentation, and its feature adaptation is based on adversarial learning.
SIFA \cite{chen2020unsupervised} and SASAN \cite{tomar2021self} are the state-of-the-art methods based on synthetic images, derived from CycleGAN framework. 
For all these baselines, we directly use the codes provided by the authors on github,
and the exact same setting is used when comparing these methods with ours, to make a fair comparison.
\emph{CISFA} (no weight) and \emph{CISFA} are both our approaches, no weight referring to no weights on $L_{pcl}$, thus having no information in the label space. 
Meanwhile, we also provide the segmentation performance of supervised training with all labels on the target domain. 
We do not need to beat this baseline as it is fully supervised, but can view it as an important reference as the performance ceiling of any UDA methods without labels.
On the other hand, w/o adaptation refers to directly applying the model trained on the source domain to target images, which serves as the performance floor. 

\subsection{Abdominal Image Domain Adaptation}
\label{sec:abd_discussion}
\subsubsection{Comparison with the State of the Art}
We switch the source and target domains for the bidirectional experiments, and Table~\ref{table1}  present dice score CT $\rightarrow$ MRI while Table~\ref{table3} shows MRI $\rightarrow$ CT results. 
We also plot the prediction masks of different methods in Fig.\ref{fig3} to visualize segmentation performance. 
The color coding for different organs in the label space is that white, yellow, orange, and red represents liver, left kidney(LK), right kidney(RK), and spleen, respectively.

We can make several important observations from quantitative and qualitative comparisons.
First of all, although there is still a gap compared with the fully supervised method, considering that we do not any target domain annotations, the gap might be reduced if using a very small amount of sparsely labeled data to fine-tune the weights in the segmenter. 
Thus, our proposed method still proves to be of high clinical practical values.
In addition, \emph{CISFA} has the highest dice score for all four organs, and increases the average dice of existing best methods by 5.98 and 2.72 percent in the two tasks. 
In terms of statistically significance, the p-values for t-test comparing CISFA with CISFA w/o weights, SIFA, and SASAN are $0.0018$,  $<0.001$ and $<0.001$, respectively, while in Table. 2, these p-values are 0.015, $<0.001$, and $<0.001$, respectively.
Secondly, VarDA fails to get segmentation performance comparable to other image synthesis approaches in the abdomen dataset.
There is a noticeable amount of false positives for liver and spleen in the visual results, which imply that only feature adaptation is not sufficient for significant domain shift cases.
Thirdly, in terms of our methods, \emph{CISFA} achieves higher average dice in segmenting target domain images than \emph{CISFA} (no weight), which shows the benefits brought by the improved $L_{pcl}$. 
Lastly,  as shown in Fig.\ref{fig3}, without any adaptation technique, the prediction mask is meaningless, while our methods output segmentations that are closest to the ground truth among all the UDA methods. All these illustrate that 
\emph{CISFA} is an effective complement to current UDA methods in medical imaging segmentation when it is difficult to get target domain annotations. 
\begin{table}[t]
    \scriptsize
    \centering
    \caption{Ablation study of two contrastive losses in CISFA on the CT $\rightarrow$ MRI adaptation task. Patch-wise contrastive is weighted.}
    \label{table5}
    \resizebox{\columnwidth}{!}{
    \begin{tabular}{ccc|cc}
    \toprule
 $L_{pcl}$ & $L_{gcl}$ & how $L_{gcl}$ is combined & Dice$\uparrow$ & ASSD$\downarrow$ \\
\hline
& &- &39.64 &7.63 \\
 \checkmark & & - & 55.76 & 3.39 \\
 & \checkmark & sum & 46.63 & 4.86 \\
 \checkmark & \checkmark & sequential & 66.50 & 3.47 \\
\checkmark & \checkmark & sum &\textbf{76.24} &\textbf{2.52} \\

\bottomrule
\end{tabular}
}
\end{table}

\subsubsection{Ablation Study}
Among the contributions of this work, the benefits of introducing weights in patch-wise contrastive loss is clear from the comparison between \emph{CISFA} (no weight) and \emph{CISFA}. 
The benefits of using contrastive loss over adversarial learning for feature adaptation is clear from the comparison between VarDA and \emph{CISFA}. Here, we further show that both the  weighted patch-wise loss $L_{pcl}$ and global contrastive loss $L_{gcl}$ are important in \emph{CISFA} by comparing the dice score and ASSD from different configurations as shown in Table~\ref{table5}.
According to the table, removing either of the two losses leads to performance degradation.
When no contrastive loss is included, \emph{CISFA} has the lowest segmentation accuracy. 
On the other hand, the influence of $L_{pcl}$ seems to be more significant than that of $L_{gcl}$. 
In our opinion, this is because $L_{pcl}$ directly determines translated image quality, and cutting $L_{pcl}$ means no shape consistency constraint.
If organ structures get distorted, it makes no sense for the global contrastive loss to draw images with different content closer in the feature space.
Notice that \emph{CISFA} without either component is different from CUT \cite{park2020contrastive}. 
The segmenter is integrated into the image synthesis flow, and there is $G_{s}$ to distinguish between prediction masks of real and fake target domain images.

We also explore the impact of how $L_{gcl}$ is combined with 
other losses in the overall workflow. 
There are actually two choices: one is to directly add the loss to all the other losses relevant to the segmenter, denoted as ``sum'' in the table; and the other is to update the weights of the encoder before optimizing other losses for the segmenter at every training iteration, denoted as ``sequential''.
It turns out that the former is better for medical imaging domain adaptation as shown in the table.
Although the logic of ``sequential'' is similar to pretraining in most self-supervised works, the discrepancy in the objectives of sequential weight update process may create problems for the training. 
After updating the weights in the encoder for minimizing $L_{gvl}$, the subsequent segmentation training also modifies them but with the purpose of reducing dice loss. 

\begin{figure}
    \centering
    \includegraphics[width=\linewidth]{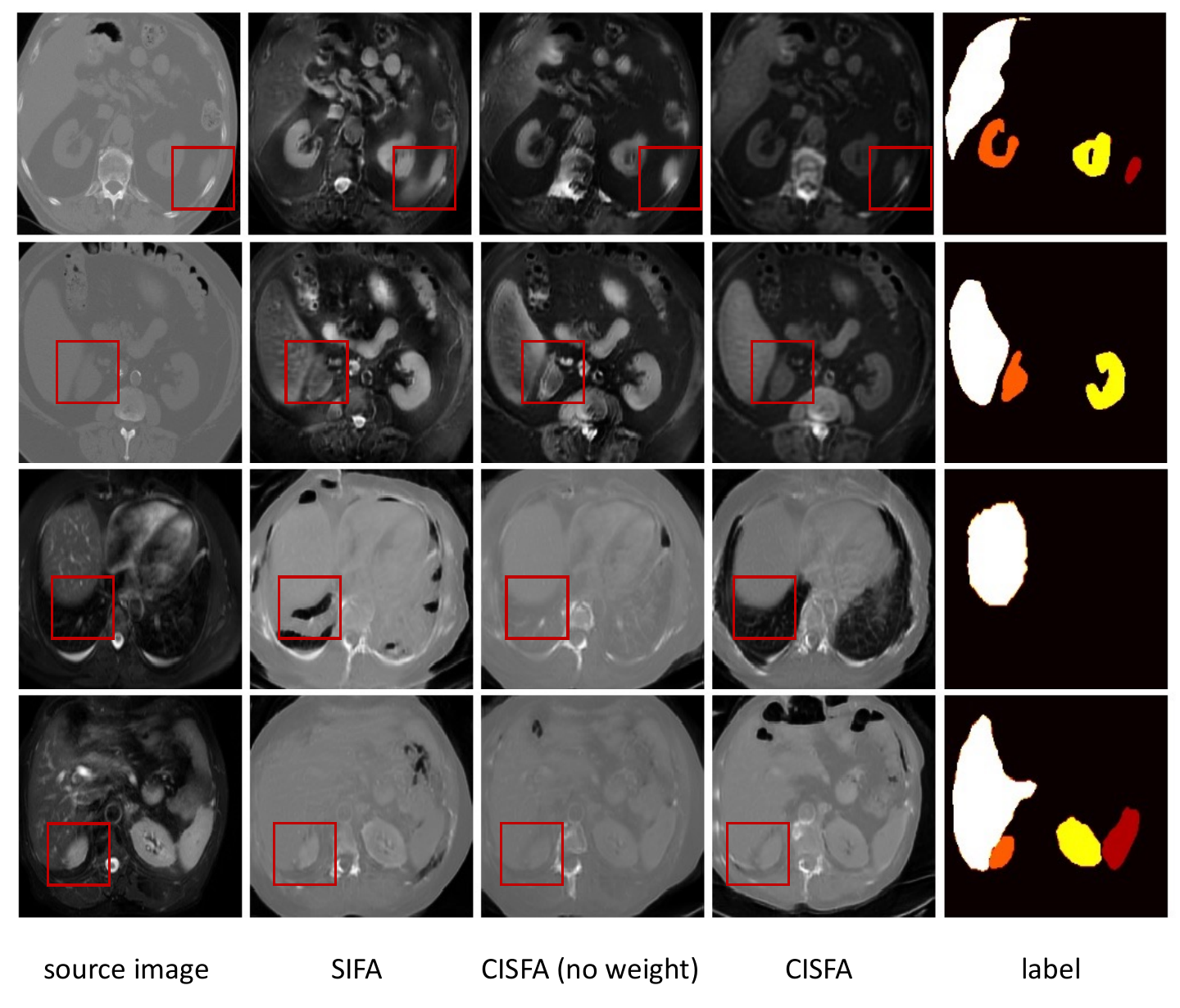}
    \caption{Images of original images and the corresponding synthetic images via SIFA, CISFA (no weight), and CISFA. The upper two rows are translating CT to MRI, and the bottom two rows are translating MRI to CT. We also attach the ground truth of each source image to show the ROI.  }
    \label{fig4}
\end{figure}

\subsubsection{Image Translation}
Fig.\ref{fig4} displays original translated images for the two different domain adaptation tasks. In general, all the methods succeed in translating the source domain images to a fake target domain that is quite similar to the real target domain, with only slight differences in brightness, contrast, and quality.
However, with a closer look at the generated images, we can observe some deformations and blurring of the organ regions of interest from SIFA.
In the first row, it is evident that the spleen (red) is expanded regarding the shape, and the boundary to background is also blurred for SIFA.
Additionally, in the third row SIFA wrongly translates the area in the red box by adding a new structure in the cavity.
These distortions are all related to areas containing subject organs, which will then influence the followup supervised segmentation training.
When it comes to comparing the two version of our proposed method, in the second and last row, the right kidney and liver areas are more obvious for \emph{CISFA}.
This phenomenon is caused by the increased weights on subject organs in our pair-wise contrastive loss, which function as an attention mechanism so that the generator addresses content in the relevant areas more.

\subsection{Whole Heart Image Domain Adaptation}
We also compare our methods with the state-of-the-art methods under the same setting on MMWHS dataset, as shown in Table~\ref{table6}.  It can be noticed that the suprvised method have higher dice score on MMWHS dataset than the abdominal dataset, which is might due to larger number of total slices for all cardiac scans. 
We can see that if we directly apply the model trained on MRI to CT dataset (w/o adaptation), the dice is lower than that in the abdominal dataset and ASSD could not even be computed because of the large number of false positives, which indicates that domain adaptation is more challenging for this task.
The reasons behind the bad performance of "W/o adaptation" have two aspects.
On one hand, it reflects that the domain shift between MRI and CT scans in the MMWHS dataset is more drastic than that in the previous abdominal dataset.
On the other hand, we observe that there is also a large variation between CT scans and it might due to being collected from different institutions and CT scanners.
\begin{table}[t]
    \scriptsize
    \centering
    \caption{Comparison between the state-of-the-art and the proposed methods on MMWHS dataset for MRI $\rightarrow$ CT adaptation.}
    \label{table6}
    \resizebox{0.8\columnwidth}{!}{
    \begin{tabular}{c|c|c}
    \toprule
 Method & Dice\% & ASSD \\
\hline
Supervised &89.78$\pm$1.26 &0.33$\pm$0.05 \\
W/o adaptation & 3.13$\pm $1.99 & - \\
\hline
CUT\cite{park2020contrastive} & 37.28$\pm$8.32 & 3.37$\pm$1.54\\
VarDA\cite{wu2021unsupervised} &40.36$\pm$2.86 &2.74$\pm$0.67 \\
SASAN\cite{tomar2021self} &61.74$\pm$3.34 &1.80$\pm$0.78 \\
SIFA \cite{chen2020unsupervised}& 64.50$\pm$4.21 & 2.14$\pm$1.21 \\ \hline
CISFA (ours) & \textbf{68.87$\pm$3.15} & \textbf{1.49$\pm$0.31} \\

\bottomrule
\end{tabular}
}
\end{table}
Therefore, the UDA experiment results have a larger gap to the supervised training performance in MMWHS than the previous dataset.
Despite that, we can draw almost the same conclusion from Table~\ref{table6} as the discussion in section ~\ref{sec:abd_discussion}.
Firstly, Feature-alignment methods, like VarDA fails to output a satisfactory segmentation accuracy in contrast to style-transfer methods, like SASAN and SIFA.
Secondly, The experiment results show that our method CISFA achieve higher dice score and lowest ASSD than other approaches on this task, which demonstrates that our proposed method can be generalized well to a different dataset. 
The p-value comparing CISFA with SASAN and SIFA are both less than 0.01. 


\section{Conclusion}
In this paper, we proposed a novel framework which builds on image domain translation and unsupervised feature adaptation for cross-modality biomedical image segmentation. 
We introduce a new weighted patch-wise contrastive loss to directly exert shape constraint on the input images and translated images, with special attention on non-background patches. 
Meanwhile, we innovatively use self-supervised representation learning as feature adaptation to improve segmentation performance. Experiments on two public datasets convinced the superiority of our method over state-of-the-art. 
In the future, we will further reduce the gap between our methods and the supervised training baselines in order to make our CISFA framework applicable to real clinical scenarios.
We might introduce a few sparsely labeled target domain images and test how much we can reduce the annotation efforts if we would like to get the equally accurate segmentation as fully supervised training, which is definitely beyond the scope of unsupervised domain adaptation.

{\small
\bibliographystyle{ieee_fullname}
\bibliography{egbib}
}

\end{document}